# A Hybrid Transformer-Sequencer approach for Age and Gender classification from in-wild facial images

(Pre-Print)

**Aakash Singh & Vivek Kumar Singh[1]**

Department of Computer Science, Banaras Hindu University, Varanasi-221005, India.

**Abstract:** The advancements in computer vision and image processing techniques have led to emergence of new application in the domain of visual surveillance, targeted advertisement, content-based searching, and human-computer interaction etc. Out of the various techniques in computer vision, face analysis, in particular, has gained much attention. Several previous studies have tried to explore different applications of facial feature processing for a variety of tasks, including age and gender classification. However, despite several previous studies having explored the problem, the age and gender classification of in-wild human faces is still far from the achieving the desired levels of accuracy required for real-world applications. This paper, therefore, attempts to bridge this gap by proposing a hybrid model that combines self-attention and BiLSTM approaches for age and gender classification problems. The proposed model's performance is compared with several state-of-the-art model proposed so far. An improvement of approximately 10% and 6% over the state-of-the-art implementations for age and gender classification, respectively, are noted for the proposed model. The proposed model is thus found to achieve superior performance and is found to provide a more generalized learning. The model can, therefore, be applied as a core classification component in various image processing and computer vision problems.

## 1. Introduction

The advancements in computer vision and image processing techniques have led to emergence of new application in the domain of visual surveillance, targeted advertisement, content-based searching, and human-computer interaction etc. Out of the various techniques in computer vision, face analysis, in particular, has gained much attention [1], [2], [3], [4]. Human face contains features that can be used to determine identity, emotions and the ethnicity etc. of people [5], [6]. The facial features and their processing have been exploited in the past in several domains ranging from security and video surveillance to customer relationship management and human-computer interaction [7], [8], [9]. Therefore, many previous studies have tried to explore different applications of facial feature processing for a variety of tasks, including age and gender classification [10], [11], [12]. Despite several previous studies having explored the problem, the age and gender classification of in-wild human faces is still far from the achieving the desired levels of accuracy required for real-world applications.

---
[1] Corresponding author. Email: vivek@bhu.ac.in

The advancement in the field of Machine learning (ML) and Deep Learning (DL) has greatly impacted the area of Computer vision [13]. The application of DL methods like convolutional neural networks, autoencoders, etc. in the area of computer vision and image classification has led to significant accomplishments [14]. Age and gender classification from facial images is one such area explored in several previous studies such as [10], [11], [12]. A broad categorization of the different approaches employed for this task is presented in **Figure 1**. Most of the early studies applied some kind of machine learning models [17-21]. However, one major weakness of such methods was that the performance of the model was quite sensitive to the feature extraction and selection mechanism. The accuracy of the models applied was also not very high in general. The advent of deep learning opened new possibilities of computations on image data. Different deep learning models have been applied on image data processing and provided the facility to automatically extract the important features [22-24]. Further, the deep learning-based models are capable of handling large training data, though sometimes that becomes a bottleneck as the availability of domain-specific annotated data is not always available. The recently introduced concept of Transfer Learning provided a better approach to handle such data availability problems [29,30]. This technique allows the knowledge gained on bigger and more general data to be fine-tuned and used on some smaller domain-specific problems. Although this technique is promising and convenient to address many modern problems, it tends to saturate at certain learning levels. To achieve aerial accuracy and reliability researchers have explored some hybrid models [31-33]. Different combinations of models (ML with DL, or multiple DL models together) are being explored. However, a carefully knit hybrid model requires both organizational knowledge of member models and structural knowledge of the problem to transcend.

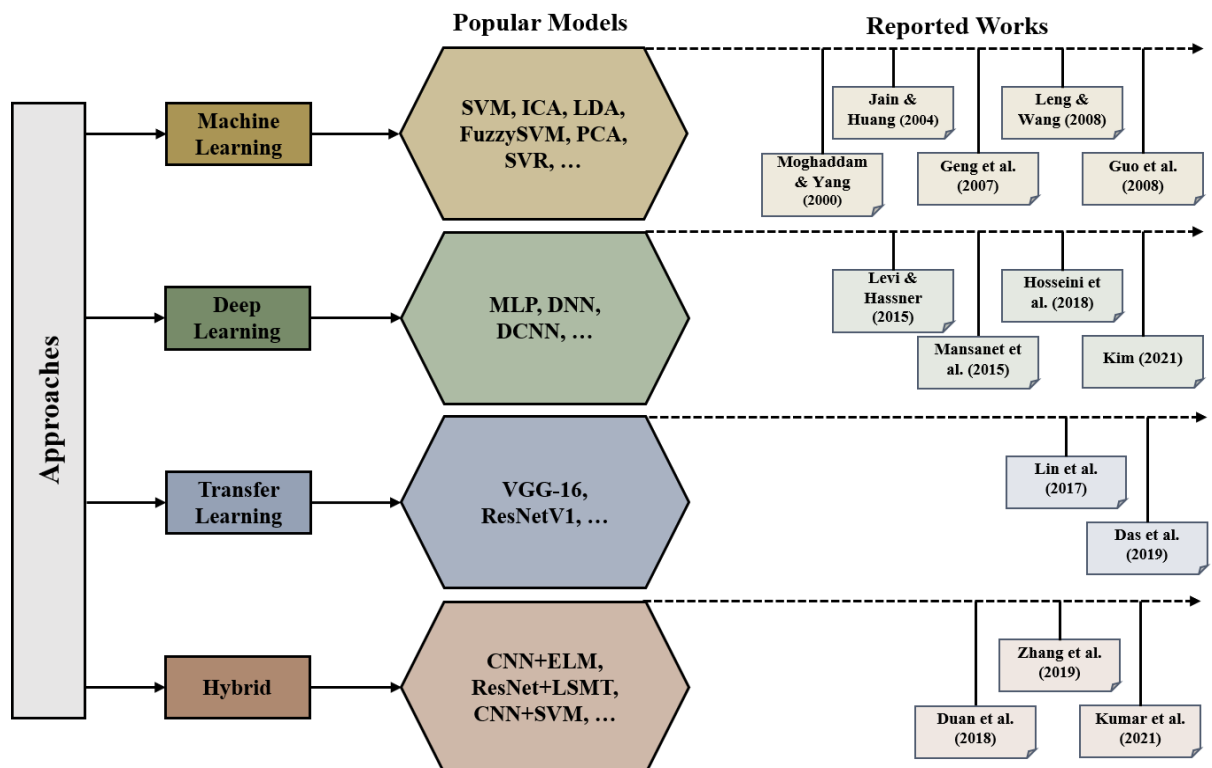

**Figure. 1:** A broad categorization of models used in Age and Gender classification problem.

Most of the recent state-of-the-art (SOTA) implementations including hybrids in the area rely heavily on the efficiency of CNN-based architectures to extract and learn the importance of various features (both low-level and high-level) for sample images. In the domain of age and gender prediction from a facial image, the accuracies reached by the various SOTAs on modern benchmark datasets are still far from the desired level for real world applications. Reliability, which includes the aspects of data resiliency and feasible training complexity, also remains the least explored in the domain. Therefore, there is a need to explore newer hybrid algorithmic approaches for the problem, which can produce more accurate and robust results. Motivated by this research gap, this work attempts to propose a novel hybrid solution for the age and gender classification problem, which can produce more accurate results as compared to various SOTAs. The recently proposed idea of self-attention mechanism [15] in Natural language processing (NLP) and its application on image-related data by [16] is explored for the problem of age and gender classification of in-wild facial images. The multi-headed self-attention mechanism of transformer (ViT) helps to weigh individual patches of an image separately, which when combined with spatial processing capabilities of deep LSTM layers, leads to an improved and generalised learning. This approach efficiently exploits the power of these two DL-based techniques to adequately characterise facial features. The high coefficients of accuracy and reliability achieved on the benchmark dataset validates the supremacy of the proposed model.

## 2. Related Work

The task of identification of Age and Gender from a given image has been studied for quite some time within the computer vision domain. The availability of several benchmark datasets has increased the researcher's capability and attention toward this problem. A tabular representation of the important previous studies in the area is given in **Table 1.** The research on this problem is seen from the year 2000 [17]. Most of the studies in the decade of 2000 have tried to address the problem with machine learning-based (ML) approaches. Among such approaches, the Support Vector Machine (SVM) [17], [18] and Liner Discriminant Analysis-based classifiers (LDA) [19], [20] were found to be very popular. Regression-based approach of estimating image age was also observed in the work of Guo et al. [21]. As we proceed in time, around 2015, a shift in methodological approach used is seen from ML-based approaches to Deep learning-based (DL) approaches [22], [23], [24], [25]. This was also a period when Artificial Intelligence (AI) got a new boost in applications with the use of DL approaches [26].

Several datasets have been used in this problem domain. At the time of initial studies, the dataset available was limited to a few data points like the FERET-gender dataset[2] which contains only 946 labelled images divided into test (236 m, 236 f) and training data (237 m, 237 f). Age dataset came at a later stage with datasets like FG-Net[3] having 1002 images of 82 different persons with ages ranging from 0-69. The horizon of the problem expanded significantly with the availability of modern benchmark datasets like Adience [27], MORPH-II [28], UTK-face[4], etc. All these datasets discussed are categorized into biological age datasets as the age mentioned in the labels are actual age of the individual. There exist another class of dataset called apparent age dataset where the ages of an individual are labelled with the average annotator's opinion. One such dataset is the Chalearn LAP[5] dataset.

---

[2] https://mivia.unisa.it/datasets/video-analysis-datasets/gender-recognition-dataset/
[3] https://paperswithcode.com/dataset/fg-net
[4] https://susanqq.github.io/UTKFace/

With the increase in data size and introduction of the ImageNet dataset and ImageNet Large Scale Visual Recognition Challenge (ILSVRC), the concept of transfer learning also attracted attention of researchers [29], [30]. Transfer learning is a concept where the knowledge gained by a model trained on a gigantic dataset employing state-of-the-art hardware can be directly transferred to a similar problem with little modification (fine-tuning). A shift towards use of hybrid approaches was also noticed around the year 2018 [31], [32], [33]. A hybrid approach combines the strengths of two or more models and mitigates weaknesses of individual models. The study [33] uses a blend of DL-based feature extractors with an ML-based customized classifier. Other studies like [31,32] try to harness the power of two DL-based networks. Despite application of a variety of models to the problem, there is still a lot of scope for improving the accuracy of the models. Despite numerous efforts, this practical problem still suffers from the below-par performance of the models in the real world. There remains a substantial gap in the path of achieving acceptable accuracy and reliability in the domain. The present work attempts to contribute to the problem domain by addressing some of the weaknesses through deployment of a suitable hybrid model that can achieve higher level of performance on the problem.

**Table 1:** Tabular representation of previous works in the area.

| Approach | Paper | Facial Attribute | Model | Dataset | Evaluation Matric | Remarks |
|---|---|---|---|---|---|---|
| ML | Moghaddam & Yang (2000) [17] | Gender | SVM classifier with RBF and cubic polynomial kernel | FERET | Average error rate over 5-fold cross-validation | Average error rate for Overall-3.38%, male-2.05%, and female-4.79% |
| | Jain & Huang (2004) [19] | Gender | Independent component analysis for feature vector extraction & Linear Discriminant Analysis for classifier | FERET | Accuracy% | Accuracy of 99.3 was obtained in test classification |
| | Geng et al. (2007) [20] | Age | They have developed an automatic estimation method AGES using LDA | FG-NET & MORPH | MAE | AGES performed better than the then known SOTA and was comparable to that of human observers |
| | Leng & Wang (2008) [18] | Gender | FuzzySVM with Gabor feature extraction and Adaboost feature selection | FERET, BUAA-IRIP, & CAS-PEAL | Accuracy% | Their classifier was better tolerant to illumination, expression, and pose |
| | Guo et al. (2008) [21] | Age | They have developed a Locally adjusted robust regressor (LARR) | FG-NET & In-house Dataset | MAE | Their LARR has shown to achieve less MAE that both age regressor SVR & classification SVM |
| DL | Levi & Hassner (2015) [22] | Age & Gender | Convolutional neural network (CNN) | Adience | Accuracy% with 5-fold cross-validation | Mean accuracy of 86.8±1.4 for gender and 50.7±5.1 for age |
| | Mansanet et al. (2015) [23] | Gender | Deep neural network with local feature extractor | LFW & Gallagher | Accuracy% with 5-fold | Accuracies of 96.25% & and 90.85% were |

---
[5] https://chalearnlap.cvc.uab.cat/

| | | | | | cross-validation | achieved on respective datasets |
|---|---|---|---|---|---|---|
| | Hosseini et al. (2018) [24] | Age & Gender | Gabor filter with Wide CNN | Adience | Accuracy% | Age & Gender accuracy was found to be 61.3% & 88.9 |
| | Kim (2021) [25] | Age & Gender | MLP with dropouts, Batch Normalization, and Skip connections | Adience, IMDB-WIKI | Accuracy% with 5-fold cross-validation | Age & gender accuracies on audience were 60.86±2.83% & 90.66±2.84%. IMDB-WIKI was used in pre-training |
| Transfer learning (DL) | Lin et al. (2017) [29] | Age | VGG-16, pre-training & transfer learning | Adience & FG-NET | Accuracy% | FG-NET was used for fine-tuning purposes, highest accuracy achieved was 57.2% |
| | Das et al. (2019) [30] | Age, Gender & Race | They have developed MTCNN that runs on FaceNet with ResNet V1 | UTK-Face, BEFA | Accuracy% | Ranked first in BEFA challenge with accuracies for Race 84.29%, Gender 93.70% & Age 71.83% |
| Hybrid | Duan et al. (2018) [31] | Age & Gender | CNN + ELM | Adience, MORPH-II | Accuracy% with 5-fold cross validation, MAE | Adience accuracies on age 52.3±5.7% & gender 88.2±1.7%, MORPH-II MAE on age was 3.44 and accuracy on gender was higher than CNN only method |
| | Zhang et al. (2019) [32] | Age | ResNet combined with LSTM (AL-ResNets) with pre-trainings | Adience, IMDB-WIKI, FG-NET, MORPH II, LAP | Accuracy% with 5-fold cross-validation | Accuracy achieved on adience dataset was 67.83±%2.98 other datasets were used for pre-training |
| | Kumar et al. (2021) [33] | Age & Gender | They have used CNN as a backbone with Gabor + SVM as a classifier | Adience, IOG, FG-NET, FEI, In-house dataset | Accuracy% | Age accuracy on Adience 74.5%, IOG 75.7% & FG-NET 92.48%, Gender accuracy on Adience 88.3%, IOG 95.1%, FEI 94.1% & In-house 91.8% |

## 3. Dataset

The dataset used in this study for training and testing is the Adience face dataset [27]. It is among one of the most popular datasets openly available to the scientific community [34]. It was released by the Open University of Israel in 2014. Being curated from unfiltered Flicker images makes it an "in the wild" dataset. It has 2 versions; one is with raw faces and the other is with faces being aligned with their 2D in-plane face alignment tool. We have used the first version in the study having a total data of 19,370 individual images labelled with their respective age-group and gender. The age label consists of 8 groups [(0-2), (4-6), (8-12), (15-20), (25-32), (38-43), (48-53), (60,100)] and gender is a collection of 3 labels [f, m, u], where 'f' stands for female, 'm' for male and 'u' for unidentified gender. There were some data points where either age or gender or both were not mentioned. We have dealt with such cases through pre-processing as described below.

## 3.1 Data Pre-processing

The pre-processing step dealt with investigating and removing erroneous data, missing data, and noise reduction. The **Algorithm 1** specifies the stepwise pre-processing steps used. A summary of images being excluded during various step is given in **Table 2**. The net available images have gone through 5 unique pre-processing steps before being fed to classification models.

**Algorithm 1:** Dataset pre-processing

```
1.  function LoadDataset
2.         return ImagePathDataframe
3.  function LoadDetctionModel()
4.         DetectionModel = OpenCV.LoadModel(ResNet10-SSD.Caffemodel)
5.         return DetectionModel
6.  function Preprocessing(ImagePathDataframe)
7.         for each Row in ImagePathDataframe
8.             if ImagePathDataframe["gender"] is not (m or f)
9.                 Drop Row
10.            if ImagePathDataframe["age"] is Null
11.                Drop Row
12.            function DetectCropFace(ImagePath)
13.                Image=OpenCV.ImageRead(ImagePath)
14.                Image = OpenCV.Resize(Image, (224,224))
15.                Blob = OpenCV.BlobFromImage(Image)
16.                Detections=DetectionModel(Blob)
17.                for Detection in Detections
18.                    if Confidence(Detection) > 0.9
19.                        X,Y,W,H = GetBoundingBox(Detection)
20.                        Face = CropImage(Image.Copy, X,Y,W,H)
21.                        function UpdateRowWithFacePaths()
22.                    else Pass
23.         return PreprocessedDataset
```

**Table 2:** Summary of the pre-processing step

| Number of images before removal | 19370 |
|---|---|
| Images with no gender found | 779 |
| Images with gender marked unidentified | 1099 |
| Images with no age found | 1252 |
| Images with no face detected | 185 |
| Images with more than one face detected | 0 |
| Total images discarded | 3315 (17.1% of original) |
| Total of images after removal | 19370 - 3315 = 16055 (82.9% of original) |

We have identified images, where gender information was either missing or was marked as unidentified (since it would not aid in the learning process) also images with no mentioned age details, were also removed. Further, we have identified and cropped faces from the image's larger background. This was done to reduce noise from the image data due to its random background. For face detection, we have employed a ResNet10-SSD[6]-based pre-trained model available to use by OpenCV. The model can detect faces in varied lighting conditions, angles of faces, and occultations. It can also detect multiple faces within an image however such a case was not observed in our dataset. A total of 3315 images were excluded from our final processed dataset.

*3.2 Data Visualisation*

In order to understand the dataset further, we present a visualization of the dataset, as it stands after the pre-processing step. These distributions are going to be an important factor during the training phase and while we try to explain the evaluation results of our models [35]. **Figure 2(a)** gives us a glimpse of the spread of the data in the 8 age-groups. As may be noted, datapoints in all the age groups are not equal. While (25,32) age group has more representative datapoints, age group (48,53) have less data points. **Figure 2(b)** presents the distribution of datapoints in the two genders, which is seen to be comparable in size for both genders.

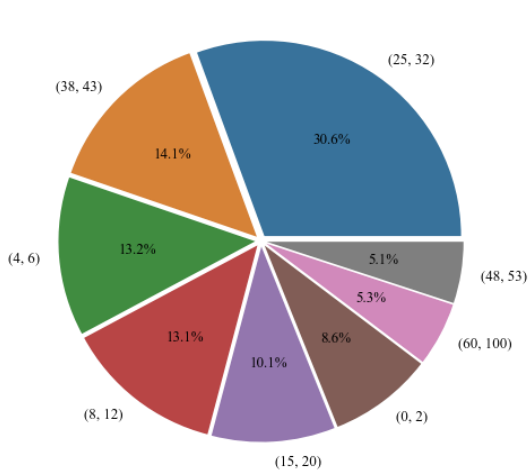 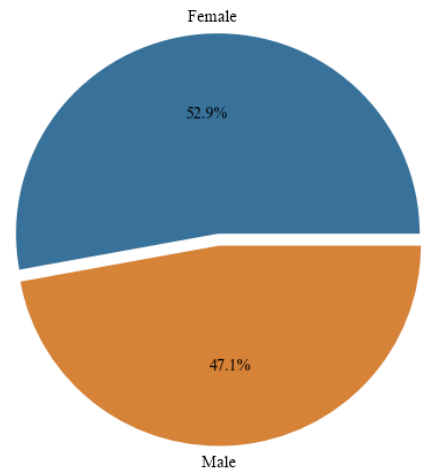

**Figure. 2(a):** Age group distribution   **Figure. 2(b):** Gender distribution

**Figure 3** presents sample images from various classes in the pre-processed dataset. It can be seen that the faces of the individual with different angles, occlusion, and lightning conditions were detected and cropped from their background to make a strictly face-only dataset.

---

[6] https://towardsdatascience.com/face-detection-models-which-to-use-and-why-d263e82c302c

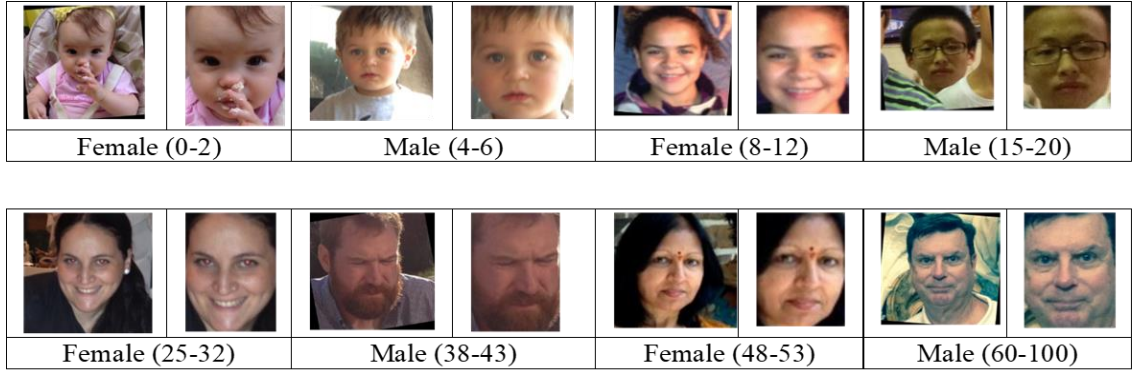

**Figure. 3**: Sample Images of different classes and their respective cropped face

## 4. Model Architecture

Traditionally most of the recent image classifiers heavily depend on CNN-based architectures, as they have been known for their versatility in feature extraction and superior performance. The introduction of self-attention-based models for image classification has created a new direction of research in the domain. Further, the work reported in [36] have shown that LSTM-based model "Sequencer" poses a good competition to self-attention-based transformer networks. Therefore, we have proposed a hybrid model that employs the intra-attention-based knowledge gained on sub-characteristics of facial attributes, which is later clubbed with spatial pattern processing capabilities of BiLSTM (hybrid-Sequencer). The model performs in an adaptive manner for the classification of images containing facial features. To validate the performance of the model, its performance is compared with two well know CNN-based models (ResNet50V2 & DenseNet121). Another comparison with a regular vision transformer (Vanilla ViT) is also performed.

*4.1 The Proposed Model*

The motivation behind the development of the model was drawn from 2 major acknowledgments about facial attributes. The first is the presence of connectedness among the facial information present in different regions of a face image. Specifically, the age and gender are the two attributes that rely heavily on this connectedness. To exploit this property, we have used ViT's multi-head self-attention that is fed with individual face image patches. Since the ViT head is pre-trained on a large data of imagenet2012 and fine-tuned on imagenet21k, it would be able to extract the connectedness between each patch of the image. The second fact is that remembering the location of age and gender-sensitive regions and focusing on that in subsequent training would improve the training process. BiLSTMs of our h-Sequencer were apt for the task. Therefore, the model is appropriately designed to exploit these facts. **Figure 4** presents a block diagram of the proposed model. Each of the components of the model are described ahead.

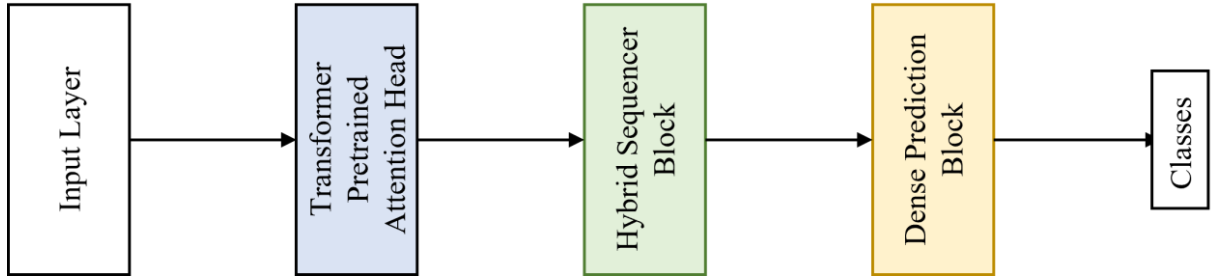

**Figure 4:** Proposed model's architecture block diagram

### 4.1.1 Transformer pre-trained attention head

The first block in the architecture diagram represents the model's input layer. It is a simple Keras input layer whose sole purpose is to define the shape of the input images and align it with the shape of the proceeding block. The next block in the diagram is of transformer's (ViT's) pre-trained multi-head self-attention. To implement this, we have used ViT's implementation in Keras (ViT-Keras) with its weights being pre-trained on imagenet21k + imagenet2012 datasets. There are several ViT models available, but we have chosen ViTB32 model. **Figure 5** shows the structure of the Transformer pre-trained Attention head. Here, B refers to base image size (224,224) and 32 refers to patch size used in the implementation. This version of ViT was selected as the other 2 CNN-based models used in the comparative study were also pre-trained on similar dimensions of images. The 2D image has to be flattened and embedded with its position patch before passing it to the transformer encoder. So an image $x \in \mathbb{R}^{H \times W \times C}$ is reshaped into a sequence of 2D flattened patches $x_p \in \mathbb{R}^{N \times (P^2 \cdot C)}$, where C is the number of channels in the image, (H, W) is the dimensions of the image, (P, P) is the dimensions of each image patch, and $N = HW/P^2$ is the effective number of patches also used as the effective length of the input sequence to the transformer. The encoder layer in the transformer contains alternating layers of multi-headed self-attention (MSA) and MLP blocks. A Layer norm (LN) is applied before and a skip connection is applied after every block [16].

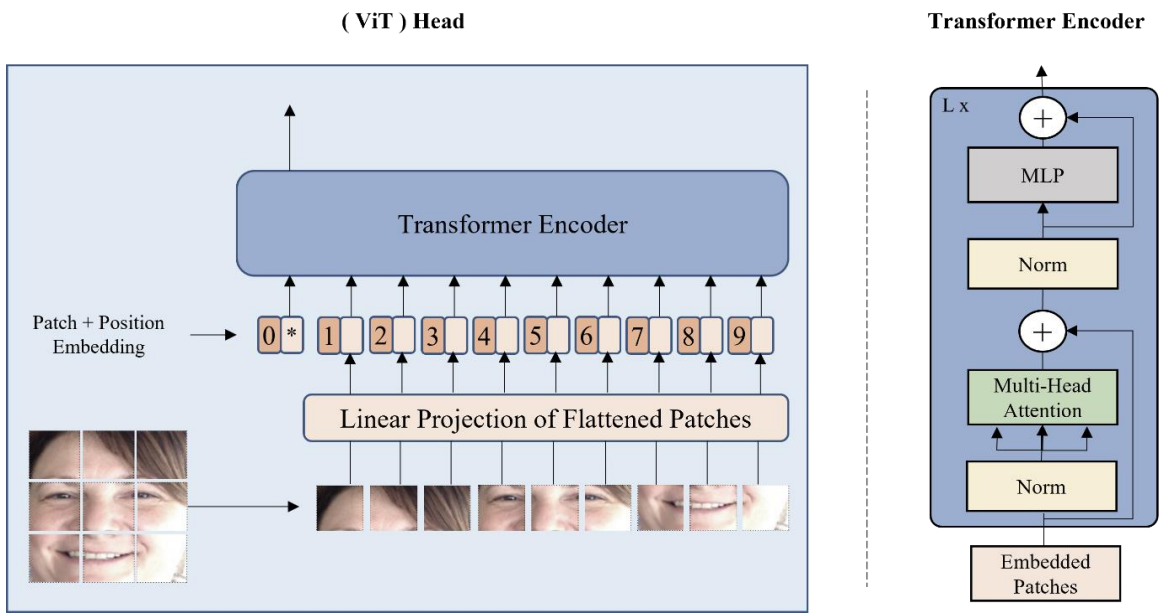

**Figure 5:** Block diagram of ViT head (pre-trained)

### 4.1.2 Hybrid sequencer block

The hybrid sequencer block comes next in the model architecture. **Figure 6** shows its structural details. This consists of 2 BiLSTMs, each preceded by a Batch normalization. First BiLSTM holds 128 nodes with *tanh* as activation function and *sigmoid* as recurrent activation function. The second BiLSTM has 64 nodes with the same activation configurations as the first. There are 3 skip connections to boost up the gradient descent process. The output is concatenated into two levels using Keras layer concatenation. The name hybrid sequencer is taken as the architecture was largely influenced by the study of [36]. This block is used in the replacement of ViT's MLP head.

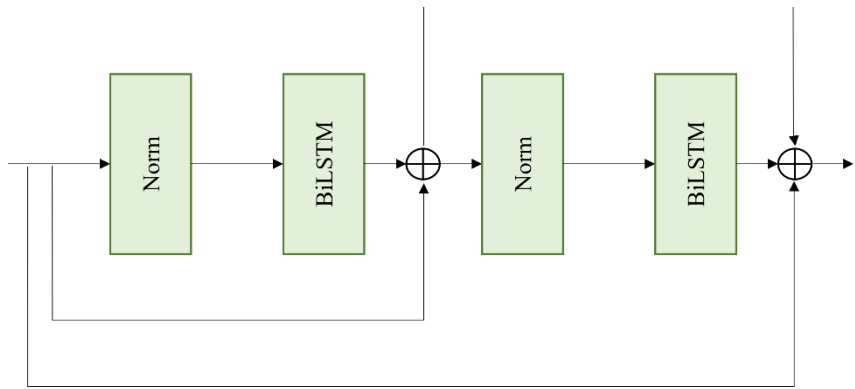

**Figure. 6:** Hybrid Sequencer Block

### 4.1.3 Dense prediction block

The last block in the model architecture is a dense prediction block. It contains only a single hidden layer and the number of neurons in that layer is defined by the number of output classes that we have to predict. Thus, there are 8 neurons for age prediction and a single neuron for gender prediction. The *softmax* activation function used for the former, and *sigmoid* activation function was used in the later.

*4.2 Other comparative models*

We have considered three other models for a comparative analysis of the performance of the proposed model. Two of these models are CNN-based models (ResNet50V2 and DenseNet121) and the third one is an attention-based model (Vanilla ViT). Since the Vanilla ViT is very much similar to the description in section 4.1 (excluding the h-Sequencer), therefore, this is not described again. Only the two CNN-based models are described next in brief.

### 4.2.1 ResNet

In 2015 Microsoft Researchers introduced a new architecture, called ResNet, that runs on the CNN backbone and utilizes the supremacy of the Residual block to cope with the problem of poor training of prevailing deep architecture models. Here the skip connections were the major advancement in this proposed network. It connects the activation of a layer by skipping a few layers in between to further layers. This makes the basic Residual block (**Figure 7**).

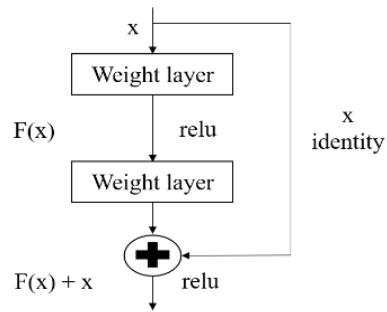

**Figure. 7:** Single residual block

To understand why it is called a residual block, let us consider a neural network block where the input is x and H(x) is the true distribution that it wants to learn. Let the difference/ residual be represented as:

$$R(x) = \text{Output - Input} = H(x) - x$$

Rearranging the terms, it gives:

$$H(x) = R(x) + x$$

These blocks enable neural networks to calculate identity functions. When these blocks are taken together and stacked up in a certain combination, it forms a Residual network [37]. For this study, we have used ResNet50V2 pre-trained model available with the Keras applications library. ResNet50V2 is an improvement on ResNet50 that is shown to perform better than ResNet50 and ResNet101 on the imagenet2012 dataset [38]. **Figure 8** presents the block diagram of the complete network (excluding the classification layer).

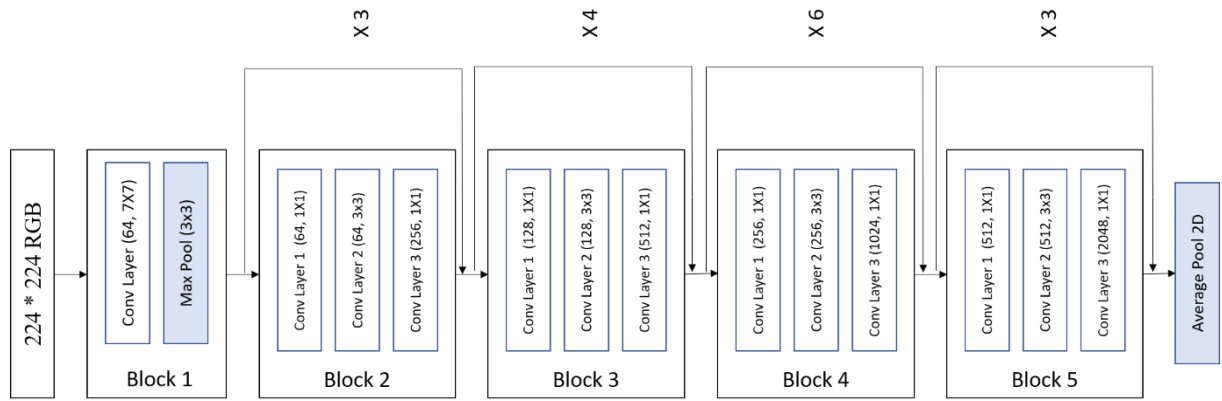

**Figure. 8:** Block diagram of ResNet50V2

### 4.2.2 DenseNet

Another popular implementation of CNN's exploiting the power of skip connection is DenseNet. It was first introduced in CVPR 2017 and received the best paper award [39]. Traditionally in feed-forward neural network, output of each layer is passed to the next layer after applying some composite of operations such as pooling, normalization, etc. It is represented by:

$$x_l = H_l(x_{l-1})$$

ResNet improved the concept by adding skip connections and changing the above equation as:

$$x_l = H_l(x_{l-1}) + x_{l-1}$$

In DenseNet each layer receives additional input in form of the feature maps from all layers preceding it. These inputs are concatenated before passing them to the next layer. The equation is reformulated as:

$$x_l = H_l([x_0, x_1, \ldots, x_{l-1}])$$

The feasibility of concatenation operation is questionable when the dimensions of the feature map change. To deal with this aspect, the concept of Dense block was introduced. A DenseNet is considered to be a combination of dense blocks, where the dimensions of the feature map do not change within the block, only the number of the filter between them is changed. There are Transition layers between two dense layers, whose role is to perform essential down samplings. For the present work, we have considered the DenseNet121 version of the architectures proposed by Huang et al. [39]. The architecture parameters of the model are displayed in **Table 3**. Pre-trained weights on the imagenet2012 dataset for DenseNet121 are available with the Keras applications library.

**Table 3:** DenseNet121 architecture with growth rate k=32

| Layers | Output Size | DenseNet-121 |
|---|---|---|
| Convolution | 112 x 112 | 7 x 7 conv, stride 2 |
| Pooling | 56 x 56 | 3 x 3 max pool, stride 2 |
| Dense Block (1) | 56 x 56 | $\begin{bmatrix} 1 \times 1\ conv \\ 3 \times 3\ conv \end{bmatrix}$ x 6 |
| Transition Layer (1) | 56 x 56 | 1 x 1 conv |
| | 28 x 28 | 2 x 2 average pool, stride 2 |
| Dense Block (2) | 28 x 28 | $\begin{bmatrix} 1 \times 1\ conv \\ 3 \times 3\ conv \end{bmatrix}$ x 12 |
| Transition Layer (2) | 28 x 28 | 1 x 1 conv |
| | 14 x 14 | 2 x 2 average pool, stride 2 |
| Dense Block (3) | 14 x 14 | $\begin{bmatrix} 1 \times 1\ conv \\ 3 \times 3\ conv \end{bmatrix}$ x 24 |
| Transition Layer (3) | 14 x 14 | 1 x 1 conv |
| | 7 x 7 | 2 x 2 average pool, stride 2 |
| Dense Block (4) | 7 x 7 | $\begin{bmatrix} 1 \times 1\ conv \\ 3 \times 3\ conv \end{bmatrix}$ x 16 |
| Pool | 1 x 1 | 7 x 7 global average pool |

5. **Experimental Setup**

Now we discuss the experimental setup, including the configurations used in the training step. **Figure 9** presents an overview of the steps in the experiment design. Nvidia A5000 RTX 24GB graphics unit was used as the hardware experimental runs. We have evaluated all the models discussed in section 4 using 5-fold cross-validation [40] in two separate ways. One is without data augmentation (DA) and the other is with data augmentation. For the latter, we have induced randomness into data by randomly flipping, transposing, saturating pixels, and rotating the images at different angles. The mean of the 5-fold evaluation of all models in both modes (without DA & with DA) is then calculated and presented in a table for a clear comparison. The validation set is taken to be 20% of data available in the training fold.

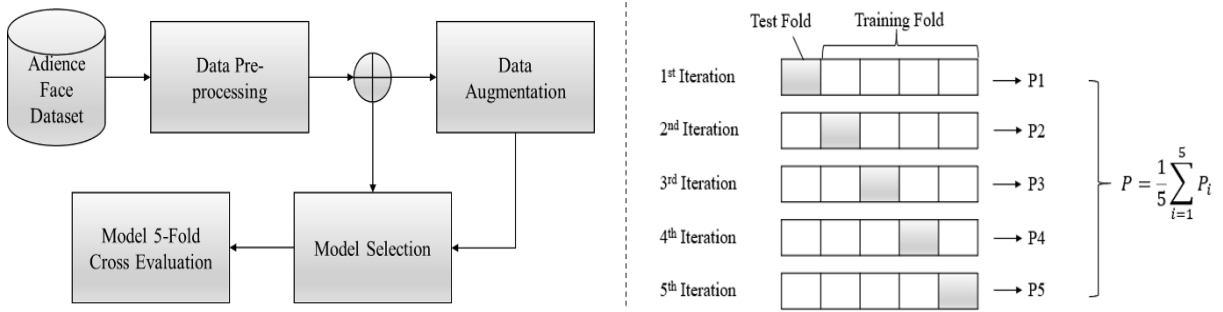

**Figure 9:** (Left) Flow-chart of the experimental setup. (Right) Cross-validation step

The configurations used in the training and verification process are detailed now. It may be noted that the configurations remain the same for all the models to facilitate a non-partial performance comparison. Keras and TensorFlow 2.0 were the major python framework used to perform the desired task. After the development of the proposed model and implementation of the other models discussed, we have applied the models to the pre-processed Adience face dataset. This process involves loading the dataset images in the form of pixel matrix clubbed together into batches of size 32. The images were resized to a shape of (224, 224, 3). Few standardization and normalization techniques namely rescaling the pixel intensity between 0-1 from 0-255, taking sample-wise center, and sample-wise standard normalization at the batch level were employed. Where the sample-wise center shifts the origin of intensity distribution by subtracting with its sample's mean $x_i = x_i - \sum \frac{xi}{n}$. The sample-wise standard normalization scale the intensity by dividing the value by its sample's standard deviation $x_i = \frac{x_i}{\sqrt{\frac{\sum |x-\bar{x}|^2}{n}}}$. This step helps to deter slow or unstable learning processes by allowing faster convergence. It also diminishes the chances of exploding gradient problems while training [41]. **Figure 10** presents a visualization of the pixel intensity distributions using histogram plots of before and after normalization.

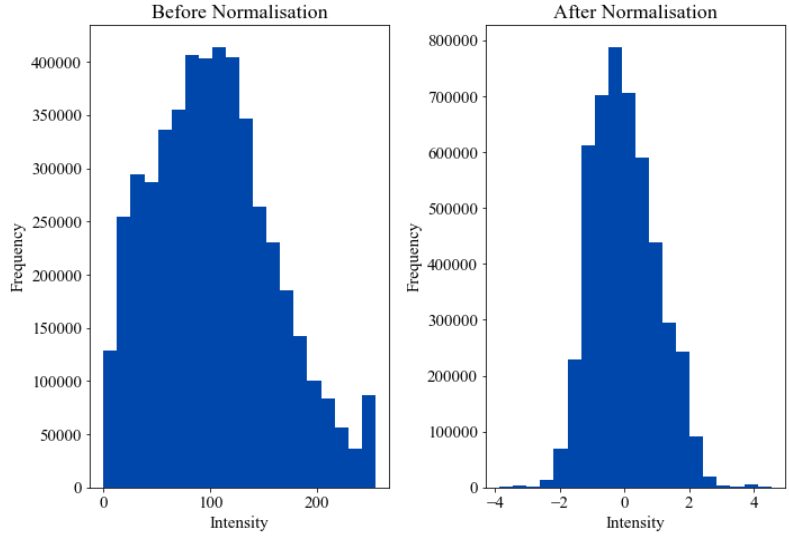

**Figure. 10:** Histogram plot of batch pixel intensities.

**Model compilation**: The loss for the models was calculated using cross-entropy loss. It calculates the difference between two probability distributions for a selected random variable or set or values [42]. Let cross-entropy be denoted as H(P,Q), where P is the target distribution and Q is approximation of target distribution for event x, then

$$H(P, Q) = -\sum_{x \in X} P(x) * \log(Q(x))$$

We have used rectified Adam (RAdam) as the optimiser with dynamic learning rates. RAdam is shown to handle the large variance problem of Adam in the early stage of training on a limited dataset. It uses a variance rectification term

$$r_t = \sqrt{\frac{(\rho_t - 4)(\rho_t - 2)\rho_\infty}{(\rho_\infty - 4)(\rho_\infty - 2)\rho_t}}$$

where $\rho_t$ is representing $f(t, \beta 2)$, $t$ represents time step, $\beta 2$ represents decay rate of moving 2$^{nd}$ moment and $f(t, \beta 2)$ represents the length of simple moving average (SMA). This helps RAdam to achieve better accuracy in a lesser number of epochs [43]. The learning rate (lr) was manipulated by ReduceLROnPlateau call-back provided by Keras. The learning starts with comparatively big lr = 1*e-4 which allows faster learning at the beginning. The lr is reduced by a factor of 0.2 (>=1*e-6) by monitoring improvements in validation accuracy at the end of an epoch. We have turned on label smoothing by the factor of 0.2. It is shown in [44] that label smoothing can help to avoid overfitting problems to an extent by allowing the model to learn more generalized features.

**Callbacks**: Callbacks are special objects provided by Keras library that are designated to perform actions at various stages of model training. In this sub-section, various callbacks used in the training phase are discussed. As discussed above, we have used ReduceLROnPlateau callback (in-built) to dynamically reduce the learning rate of the model during intermediate training stages. The EarlyStopping callback (in-built) was employed to stop the training when the validation accuracy metric stops improving for certain (n=5) consecutive epochs. Once the condition is met the callback stops the training process and restores the values of the weights learned at the epoch having the highest validation accuracy. This helped us to cut down the training time by reducing the total number of epochs and also eliminating any chances of model overfitting. To boost the robustness of model training a ModelCheckpoint (in-built) was put to use. The callback saves the best model weights into a physical.hdf5 file in the working directory during the training phase. In the event of unexpected system failure, we may load the saved .hdf5 file and resume training from the point where it was interrupted. The next GC callback was a costume callback designed to handle garbage collection and memory leak problems during intermediate epochs. At the end, one more costume callback named TimeCallback was used to store down the training variables like epoch execution time, and the total number of epochs during each cross-fold validation stage.

**Evaluation Metric:** The evaluation metric for a model gives an indication of model performance. Accuracy was used as a metric in our case as it is a widely used metric. For this purpose, confusion matrix was made and then using these values, Accuracy was computed as follows:

$$Accuracy = \frac{TN+TP}{TN+FP+TP+FN}.$$

where, TP- True Positive, TN- True negative, FP- false Positive and FN- False negative.

## 6. Results

This section presents the experimental results obtained on both the problem tasks- age classification and gender classification.

### 6.1 Age Classification

The age classification was a multi-class classification problem. The Adience Face Dataset has 8 defined age-groups. The proposed model and the other comparative models were applied on the dataset. The confusion matrix for the proposed model is shown in **Figure 11**.

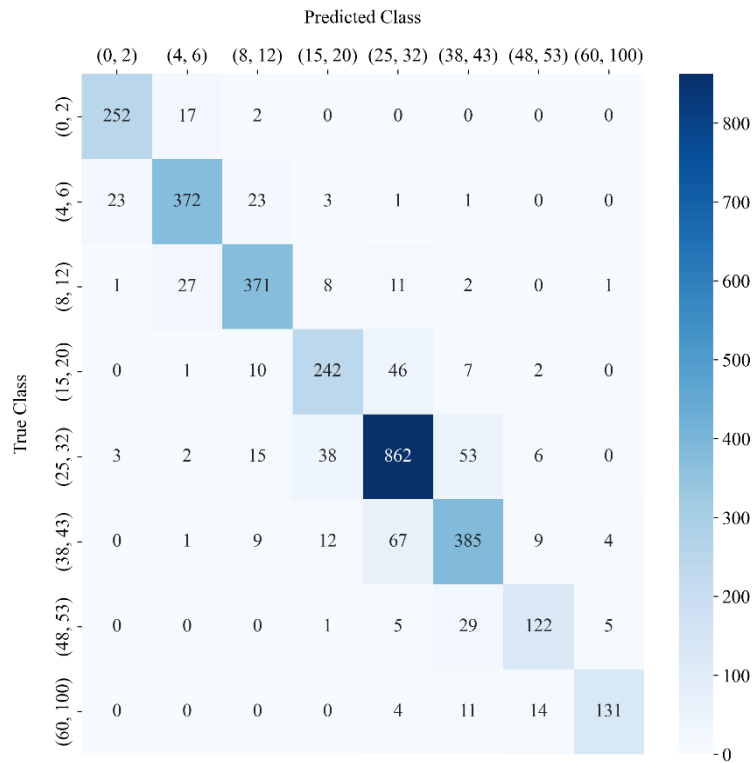

**Figure. 11:** Age classification confusion matrix

From the confusion matrix, it can be seen that the proposed modal has performed quite well in predicting the age-groups. Further, the confusions are largely found to lie next to one another, indicating that the model would be useful in real world settings, where an approximate classification may be acceptable. The proposed model obtained a test accuracy of 84.91 $\pm 0.84\%$, which is significantly higher than other models discussed and also higher than any SOTA reported till date. **Table 4** presents the F1 scores for the individual classes. The class receiving the least F1 score was observed to be (48,53). A probable justification for this may be the fact that this class lacked enough training samples.

**Table 4:** F1 scores for all age-groups in the test dataset

| Class | (0,2) | (4,6) | (8,12) | (15,20) | (25,32) | (38,43) | (48,53) | (60,100) |
|---|---|---|---|---|---|---|---|---|
| F1 Score | 0.92 | 0.88 | 0.87 | 0.79 | 0.87 | 0.79 | 0.77 | 0.87 |

The loss and accuracy curves of the training phase are shown in **Figure 12(a)** and **Figure 12(b)**, respectively. Both depict a smooth convergence without any over-fitting.

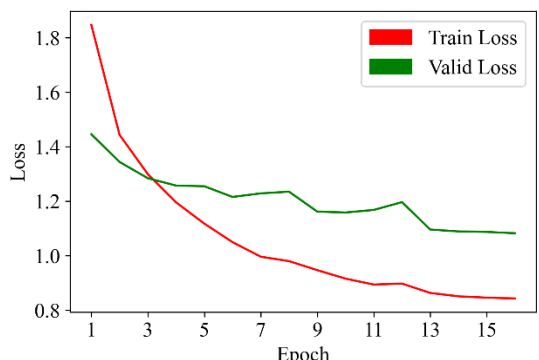
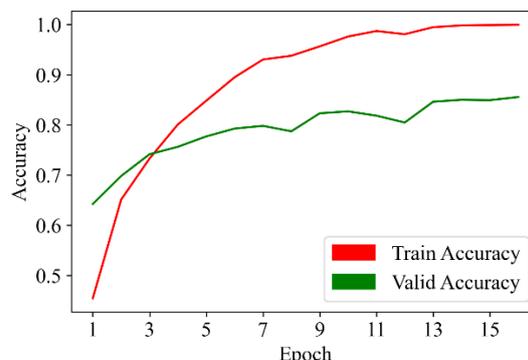

**Figure 12(a):** Age classification loss     **Figure 12(b):** Age classification accuracy

The performance of the proposed model is also compared with results obtained from other comparative models. **Table 5** provides a comparative analysis of the various parameter values of the different models for the age classification task. The table contains the mean values of performances shown in 5-fold cross-validation along with their standard deviation values. An extra detail of "epoch time mean" is given in the last column of the table to understand the time complexity of each model in the training phase. The proposed model shows a test accuracy of 84.91%, which is significantly higher as compared to other models implemented, and approx. 10% more than the SOTA. It may be noted that the self-attention-based models are not much affected by the induction of randomness through Data Augmentation (DA). They are also seen to show far less deviation from the mean accuracy when compared with CNN-based models. Thus, from the table, one may note that the proposed model was as resilient as a vanilla self-attention model (ViT) with increasing the accuracy metric. Therefore, it may be said that the model was able to learn more accurate and generalized information about the dataset without largely affecting the training complexity.

**Table 5:** The loss and accuracy of the trained models.

| Age classification models | Training loss mean (Std) | Training accuracy mean (Std) | Validation loss mean (Std) | Validation accuracy mean (Std) | Test loss mean (Std) | Test accuracy mean (Std) | Epoch Time Mean in Sec (Std) |
|---|---|---|---|---|---|---|---|
| **ResNet50v2** | 0.8138 (0.0032) | 0.9999 (0.0001) | 1.2122 (0.0362) | 0.7912 (0.0135) | 1.2172 (0.0315) | 0.7846 (0.0168) | 62.3 (0.22) |
| **ResNet50v2 (DA)** | 0.8762 (0.0184) | 0.9887 (0.0046) | 1.2127 (0.0160) | 0.7817 (0.0061) | 1.2041 (0.0134) | 0.7874 (0.0069) | 130.2 (5.16) |
| **DenseNet121** | 0.8233 (0.0144) | 0.9999 (0.0001) | 1.4031 (0.1108) | 0.6857 (0.0765) | 1.4077 (0.1064) | 0.6845 (0.0740) | 100.6 (0.34) |
| **DenseNet121 (DA)** | 0.9596 (0.0360) | 0.9662 (0.0184) | 1.3055 (0.0200) | 0.7413 (0.0101) | 1.3084 (0.0352) | 0.7385 (0.0101) | 159.0 (0.64) |
| **ViT** | 0.8413 (0.0232) | 0.9983 (0.0009) | 1.1425 (0.0232) | 0.8362 (0.0136) | 1.1571 (0.0051) | 0.8269 (0.0073) | 90.4 (0.81) |
| **ViT (DA)** | 0.8622 (0.0404) | 0.9886 (0.0113) | 1.1563 (0.0356) | 0.8303 (0.0118) | 1.1772 (0.0330) | 0.8148 (0.0097) | 142.7 (0.86) |
| **SOTA** [33] | | | | | | 0.7450 | |
| **ViT-hSeq** | 0.8345 (0.0053) | 0.9997 (0.0001) | 1.0876 (0.0220) | 0.8533 (0.0107) | 1.0907 (0.0156) | **0.8491** (0.0084) | 88.4 (1.44) |
| **ViT-hSeq (DA)** | 0.8513 (0.0100) | 0.9993 (0.0004) | 1.1182 (0.0041) | 0.8432 (0.0037) | 1.1297 (0.0151) | 0.8348 (0.0076) | 146.0 (1.40) |

## *6.2 Gender Classification*

The gender classification is a binary classification task as there are only two genders considered (Female & Male). The proposed as well as other comparative models were implemented, and the various output parameters were observed. **Figure 13** presents the confusion matrix for gender classification. There are few misclassifications in test predictions. The proposed model obtained an accuracy of 96.56 $\pm$0.27%, which is well above any SOTA observed till date. The F1 scores of test prediction for both classes are listed in **Table 6**.

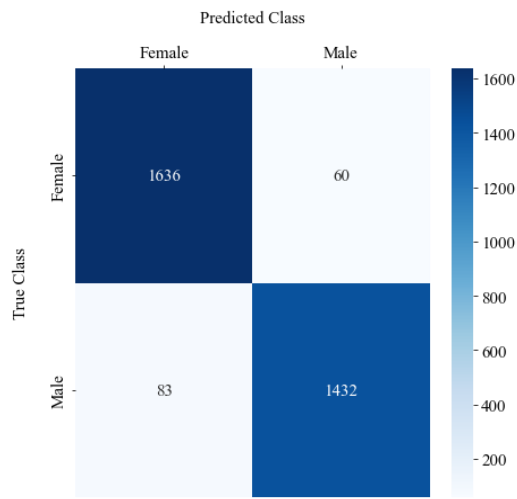

**Figure. 13:** Gender classification confusion matrix

**Table 6:** F1 scores for gender classes in the test dataset

| Class | Female | Male |
|---|---|---|
| **F1 Score** | 0.96 | 0.95 |

The loss and accuracy curves of the training phase are shown in **Figure 14(a)** and **Figure 14(b)**, respectively. Both are found to have a smooth convergence without any over-fitting.

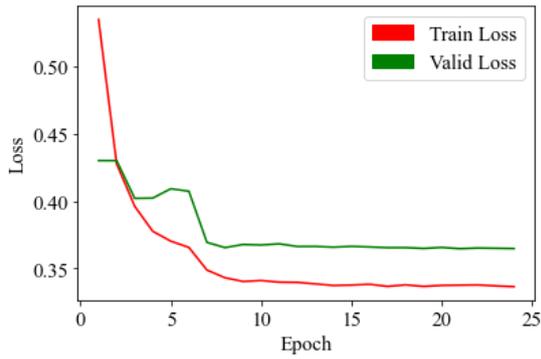
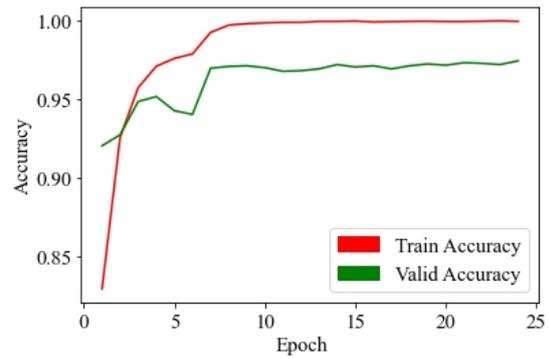

**Figure. 14(a):** Gender classification loss     **Figure. 14(b):** Gender classification accuracy

The performance of the proposed model for the gender classification task was compared with various other models implemented. **Table 7** presents the aggregate performance indicators of various models implemented. The performance indicated by these test accuracies shows the superiority of the proposed model, when compared with other models, including the SOTA (6% higher than SOTA). It can be further seen that inducing randomness in the data does not much affect the accuracies of self-attention-based models. Specifically, the proposed model inherited the characteristic along with improving the accuracy metric. Thus, it may be reiterated that the proposed model's learning is more accurate and generalized in nature.

**Table 7:** The loss and accuracy of the trained models.

| Age classification models | Training loss mean (Std) | Training accuracy mean (Std) | Validation loss mean (Std) | Validation accuracy mean (Std) | Test loss mean (Std) | Test accuracy mean (Std) | Epoch Time Mean (Std) |
|---|---|---|---|---|---|---|---|
| **ResNet50v2** | 0.3292 (0.0025) | 0.9999 (0.0001) | 0.4285 (0.0225) | 0.9202 (0.0165) | 0.4224 (0.0228) | 0.9265 (0.0156) | 63.8 (0.33) |
| **ResNet50v2 (DA)** | 0.3507 (0.0139) | 0.9906 (0.0090) | 0.4162 (0.0156) | 0.9342 (0.0107) | 0.4165 (0.0177) | 0.9323 (0.0166) | 128.1 (2.29) |
| **DenseNet121** | 0.3303 (0.0030) | 0.9999 (0.0001) | 0.4420 (0.0171) | 0.9121 (0.0247) | 0.4370 (0.0229) | 0.9164 (0.0247) | 100.7 (0.29) |
| **DenseNet121 (DA)** | 0.3491 (0.0056) | 0.9965 (0.0021) | 0.4225 (0.0139) | 0.9333 (0.0093) | 0.4236 (0.0139) | 0.9312 (0.0109) | 159.1 (2.48) |
| **ViT** | 0.3847 (0.0652) | 0.9985 (0.0003) | 0.4175 (0.0541) | 0.9651 (0.0037) | 0.4188 (0.0550) | 0.9610 (0.0030) | 83.5 (0.44) |
| **ViT (DA)** | 0.4016 (0.0506) | 0.9936 (0.0028) | 0.4326 (0.0405) | 0.9582 (0.0044) | 0.4339 (0.0438) | 0.9539 (0.0054) | 148.6 (1.99) |
| **SOTA** [25] | | | | | | 0.9066 | |
| **ViT-hSeq** | 0.3289 (0.0013) | 0.9996 (0.0002) | 0.3677 (0.0028) | 0.9697 (0.0039) | 0.3731 (0.0037) | **0.9656** (0.0027) | 87.3 (0.42) |
| **ViT-hSeq (DA)** | 0.3487 (0.0118) | 0.9863 (0.0085) | 0.3864 (0.0063) | 0.9571 (0.0062) | 0.3869 (0.0066) | 0.9542 (0.0054) | 146.8 (2.76) |

## 7. Conclusion

The paper presents a hybrid model that combines self-attention and BiLSTM approaches for age and gender classification problems. While age is a multi-class classification problem, the gender classification is a binary classification problem. The proposed model's performance is compared with several state-of-the-art models proposed so far. The proposed model is found to achieve superior performance than all previous models, including the last known state-of-the-art model. There is an improvement of approximately 10% and 6% over the state-of-the-art implementations for age and gender classification, respectively. The results further indicate that the proposed model is less affected by the input sequence and hence provides a more generalized learning. Thus, the paper presents an architecture and implementation of a new hybrid model that achieves significantly higher performance than other models. The model is demonstrated to be accurate and input resilient solution for the given facial image classification problem. Though model's superiority is demonstrated for the problems of age and gender classification, it can be used in various other image classification problems. The model can thus serve as a core classification component in various image processing and computer vision problems. As a future work, the efficacy of the proposed model in age and gender classification can be explored on datasets primarily containing occlusive faces (faces with mask, scarf, sunglasses, etc.). It may also be interesting to observe its performance in other domains relating to other facial characteristics like pose, race, and emotion estimation.

## Acknowledgments

**Conflicts of interests/Competing interests:** The authors declare that the manuscript complies with ethical standards of the journal and there is no conflict of interests whatsoever.

**Data Availability:** The data used for this research work is from a publicly available source. Code and data for experiment will be shared on reasonable request.

**Funding**: This work is supported by the extramural research grant no: 3(9)/2021-EG-II from Ministry of Electronics & Information Technology (MeITY), Government of India, and by HPE Aruba Centre for Research in Information Systems at BHU (No. M-22-69 of BHU).